\documentclass[conference]{IEEEtran}
\IEEEoverridecommandlockouts
\usepackage{cite}
\usepackage{amsmath,amssymb,amsfonts}
\usepackage{algorithmic}
\usepackage{graphicx}
\usepackage{textcomp}
\usepackage{xcolor}
\def\BibTeX{{\rm B\kern-.05em{\sc i\kern-.025em b}\kern-.08em
    T\kern-.1667em\lower.7ex\hbox{E}\kern-.125emX}}
\begin{document}

\title{Agents for Automated User Experience Testing\\
\thanks{This research was supported by FCT-Portugal through grants UID/CEC/ 50021/2020, 2020.05865.BD and by the EU H2020 RIA project iV4xr : 856716.
}
\thanks{\textbf{DOI 10.1109/ICSTW52544.2021.00049}}
\thanks{\textbf{Accepted to the AIST workshop at the 2021 IEEE International Conference on Software Testing, Verification and Validation (ICST)}}
}

\author{\IEEEauthorblockN{Pedro M. Fernandes}
\IEEEauthorblockA{\textit{INESC-ID and Instituto Superior Técnico, Univ. de Lisboa} \\
Lisbon, Portugal \\
pedro.miguel.rocha.fernandes@ist.utl.pt}
\and
\IEEEauthorblockN{Manuel Lopes}
\IEEEauthorblockA{\textit{INESC-ID and Instituto Superior Técnico, Univ. de Lisboa} \\
Lisbon, Portugal \\
manuel.lopes@tecnico.ulisboa.pt}
\and
\IEEEauthorblockN{Rui Prada}
\IEEEauthorblockA{\textit{INESC-ID and Instituto Superior Técnico, Univ. de Lisboa} \\
Lisbon, Portugal \\
rui.prada@tecnico.ulisboa.pt}
}

\maketitle

\begin{abstract}
  The automation of functional testing in software has allowed developers to continuously check for negative impacts on functionality throughout the iterative phases of development. This is not the case for User eXperience (UX), which has hitherto relied almost exclusively on testing with real users. User testing is a slow endeavour that can become a bottleneck for development of interactive systems. To address this problem, we here propose an agent based approach for automatic UX testing. We develop agents with basic problem solving skills and a core affect model, allowing us to model an artificial affective state as they traverse different levels of a game. Although this research is still at a primordial state, we believe the results here presented make a strong case for the use of intelligent agents endowed with affective computing models for automating UX testing.
\end{abstract}


\section{Introduction}
\label{sec:introduction}

Hitherto, User eXperience (UX) testing has mostly relied on real users interacting with the system under test, following a number of different approaches \cite{vermeeren2010user,rivero2017systematic}. When applied correctly, such methods allow us to gain a wealth of knowledge about the caveats of the system and take action in order to improve its UX.

Testing with users, however, is a slow endeavour. It is not usually viable to have a user testing session at each step of the iterative development process. Whereas today is common practice to run automated functionality tests each time a system suffers an alteration, the same cannot be currently done concerning UX. 

In this paper, we will propose using intelligent agents endowed with an affective model to run automated UX tests. We do not defend such agents could, in the foreseeable future, completely replace testing with real users. What we propose is that agents could be used to maintain a focus on UX throughout the entire development process, running alongside automatic functional tests.

Our research objective is to understand if such testing agents could be a viable approach to partially automate UX testing. With this intent, we have developed an agent endowed with an internal core affect model \cite{russell2003core}. This agent was then used to gain insights on the UX of different maps of a game (Sec.~\ref{sec:test_case}). The results here presented aim to be a proof of concept, showing some of the information one could obtain using UX testing agents. We believe these primordial results make a strong claim for the usefulness of such agents. We further propose a number of ways on which these agents could be improved (Sec.~\ref{sec:discussion}). 

This paper is organised as follows. In Sec.~\ref{sec:related_work} we give a brief overview of previous research in the area. In Sec.~\ref{sec:test_case} the testing environment is presented. In Sec.~\ref{sec:agent_model} we describe the core architecture of the agent. The results are described in Sec.~\ref{sec:results} and the conclusions on Sec.\ref{sec:conclusions}. Finally, in Sec.~\ref{sec:discussion} we discuss the results and propose a number of ways in which the presented UX testing agents could be improved.

\section{Related Work}
\label{sec:related_work}

Autonomous agents have been previously endowed with emotional models and agents have been used for testing software. However, to the knowledge of the authors, both these things have never been done simultaneously in order to test UX. We believe this represents a research gap.

With regards to emotional agents, two reviews have been done, providing a wealth of information on different methods and approaches for creating agents endowed with artificial emotions \cite{rumbell2012emotions,kowalczuk2016computational}. Most of these emotional agents were developed in order to be able to have more realistic interactions with users. They were therefore created to improve UX, but not to test it.

Playtesting agents have been developed that strive to play games in a similar fashion to real users. Designed to have some of the same limitations that human players have, these agents can have, for example, a field of view and a limited short-term memory. Such agents can then be used to play games and find problems and exploits that would normally require testing with real users to find. Stahlke et al. proposed a framework to create this type of agents, following heuristics based on human behaviours, like exploration, hazard avoidance and aggression \cite{stahlke2019artificial}. 

Playtesting agents have also been used to aid design. Holmgård et al. \cite{holmgard2018automated} developed agents with a range of different personas, that is, following different behaviours and with different objectives. These agents were then used to aid developers create maps for a 2D game. These agents allowed developers to predict how different players would traverse the map and make design decisions accordingly. The creation of playtesting agents has also been used to test board games \cite{guerrero2018using}, card games \cite{garcia2018automated} and frameworks for the development of playtesting agents have been proposed \cite{borovikov2019winning,pfau2017automated}.

Outside the realm of games and focusing on the problem of automated UX testing, there is research in automatic Graphical User Interface (GUI) testing \cite{vos2015testar,memon2000automated} and automatic usability testing \cite{bellamy2011deploying,katsanos2013klm}. Both these areas are of relevance to UX testing, but we were unable to find any work in them that attempted to do any real-time estimates as an agent interacted with a system. In the area of User Modelling, real-time estimates of the internal state of users is often done \cite{desmarais2012review} and some works have even focused on modelling the internal emotional states of users\cite{martinho1999cognitive,conati2009empirically}, but to out knowledge, none has done so with the intent of UX testing.



\section{Test Case: Lab Recruits Game}
\label{sec:test_case}

The system under test for this paper will be a game called Lab Recruits\footnote{https://github.com/iv4xr-project/labrecruits}. Lab Recruits is a simple 3D game where the player must interact with objects in order to achieve a goal. The only actions the player can do is move or try to interact with objects. Objects with which the player can interact will henceforth be called \textbf{interactables}.

For the examples presented in this paper, the following objects were used:

\begin{itemize}
    \item \textbf{Door with Button (relevant object):} A door that the player can open by interacting with the corresponding button. 
    This object pair is represented in game by a door and a button connected by a wire.
    \item \textbf{Simple Button (irrelevant object):} A button that is not connected to anything. Even though the player can interact with it, such an interaction is not necessary for completing the player's objective. 
    This object is represented in game by a sphere that can have different colours and which does not have a wire connecting it to anything.
    \item \textbf{Chair (the objective):} 
    Finding this chair is, in all our examples, the final objective of the player. 
    This object is represented in game by a black office chair.
\end{itemize}

Further ahead in the paper, four Lab Recruits maps will be introduced (Fig.~\ref{fig:Map1}a, \ref{fig:Map2}a, \ref{fig:Map3}a and \ref{fig:Map4}a). All of those maps follow the same basic premise: the player spawns in a maze, which it must traverse in order to find a chair. To do so, the player might have to interact with certain buttons to open doors. Finding the chair is the ultimate goal of the player and when the chair is found, the game level ends.

\section{Agent Model}
\label{sec:agent_model}

In order to be used for the UX testing of our Lab Recruits maps, the agents had to fulfil two main requirements: (a) be able to traverse the maze; and (b) record information relevant for UX assessment as they do so. Our approach to solving (a) is described in Sec.~\ref{sec:search_algorithm} and our solution for (b) in Sec.~\ref{sec:affective_critic}.

\subsection{Search and Traverse Algorithm}
\label{sec:search_algorithm}

Here he will briefly describe the AI capabilities of the agents in order to give the reader an understanding of the agent's behaviour. Our goal was to have an agent that would behave in a similar fashion to a real user.

The agent has a field of view and cannot perceive what is behind a wall or further than a specific distance. Being endowed with a spatial memory, the agent creates an internal map recording all the locations and objects it has already found. This internal map is that which the agent uses for navigation. 

The moment the agent spawns in a map, it only knows the location of that which it can directly perceive and knows which object it is trying to find: a chair. It also has the prior knowledge of how to open a door which is connected to a button. As the focus of our simulations was to test UX, it was counter productive to have the agent learn something that the grand majority of real users would already know how to do.

The agent search algorithm runs as follows:

\begin{enumerate}
    \item The agent perceives the environment and adds to its internal map all the locations and objects it has found.
    \item If the chair was found, the simulation ends.
    \item If the chair was not found but a button that opens a door was found, the agent will move towards the button and interact with it in order to open the door. If more than one door opening button was found, the agent will randomly choose one to interact with.
    \item If neither the chair nor a door opening button were found, the agent moves to the closest information limit of its internal map, striving to find more locations and objects. A location is considered an information limit if the agent does not have any information of what is beyond it. The agent considers walls hard limits and will not attempt to explore past them.
\end{enumerate}

These steps are repeated in this order until the chair is found or all off the map is completely explored.

\subsection{Affective Model}
\label{sec:affective_critic}

The ISO 9241-210:2019 defines UX as the ``user’s perceptions and responses that result from the use and/or anticipated use of a system, product or service.'', further clarifying that ``Users’ perceptions and responses include the users’ emotions, beliefs, preferences, perceptions, comfort, behaviours, and accomplishments that occur before, during and after use.'' \cite{dis20109241}.

Our approach is to endow the testing agents with an affective state and use that agents' internal state to infer properties of the expected UX.


Several theoretical models of emotion have been proposed \cite{russell1977evidence,ortony1990cognitive,plutchik1980general}, a number of which have already been used to give agents artificial emotions \cite{rumbell2012emotions,kowalczuk2016computational}. We decided to base our model on the Core Affect theory of emotions \cite{russell2003core}. This theory defends that emotions are constructed from an initial affective state through processes like attribution and appraisal. This affective state is defined using two dimensions, Valence (positive/negative connotation) and Arousal (calm/exciting).

An affective state, by itself, is not enough to define an emotional state. But according to the Core Affect theory, it is the starting point of emotions, and as such will be the starting point of our research. In Sec.\ref{sec:discussion} we will discuss how this approach could be built upon in order to characterise complete emotional states.

To use this model, we need to define how the agent's interaction with the environment will alter those affective dimensions (the appraisal rules). At this stage of the research, our main focus is understanding the feasibility of an agent based approach to automatically test UX. As such, on defining how the agent's interaction with the environment alters its artificial affective state, we decided to use simple rules. There is a lot of room for improvement over this rule based approach and some proposals will be made in Sec.\ref{sec:discussion}. 

Both the affective dimensions will be in the range $\{-5,5\}$. When the agent spawns in the environment, both dimensions are neutral, that is, having a value of 0. From that moment on, the affective dimensions are thus calculated:

\begin{itemize}
    \item \textbf{Valence:} 
    \begin{itemize}
        \item Whenever the agent is able to accomplish his objective (finding the chair) or intermediate sub-objectives (opening doors), the Valence dimension \textbf{increases} by 1.
        \item If the agent does not accomplish any objective or sub-objective for 10 seconds, the Valence dimension \textbf{decreases} by 0.4.
    \end{itemize}
    \item \textbf{Arousal:}
    \begin{itemize}
        \item Whenever the agent finds a new interactable object (buttons and doors), the Arousal dimension \textbf{increases} by 1
        \item If the agent doe not find any new interactable for 10 seconds, the Arousal dimension \textbf{decreases} by 0.4.
    \end{itemize}
\end{itemize}

The values here used are not yet an accurate representation of the affective response of a real user. To be so, they would need to be experimentally tested, which they have not yet been at this point in the research. As it stands, we wish to understand if such a simple affective model could already provide meaningful UX insights when used to test a number of different scenarios. One of the next steps on the development of the model will be to define a more robust system for the affective state update.

\section{Results}
\label{sec:results}

On this section, we describe tests with four different Lab Recruits maps. For each, we will present the changes to the affective state of the aforedescribed agent (Sec.~\ref{sec:agent_model}) as the map is traversed. Such changes will be shown both in function of the location of the agent and in function of time.

\begin{figure}[!htb]
\includegraphics[width=\linewidth]{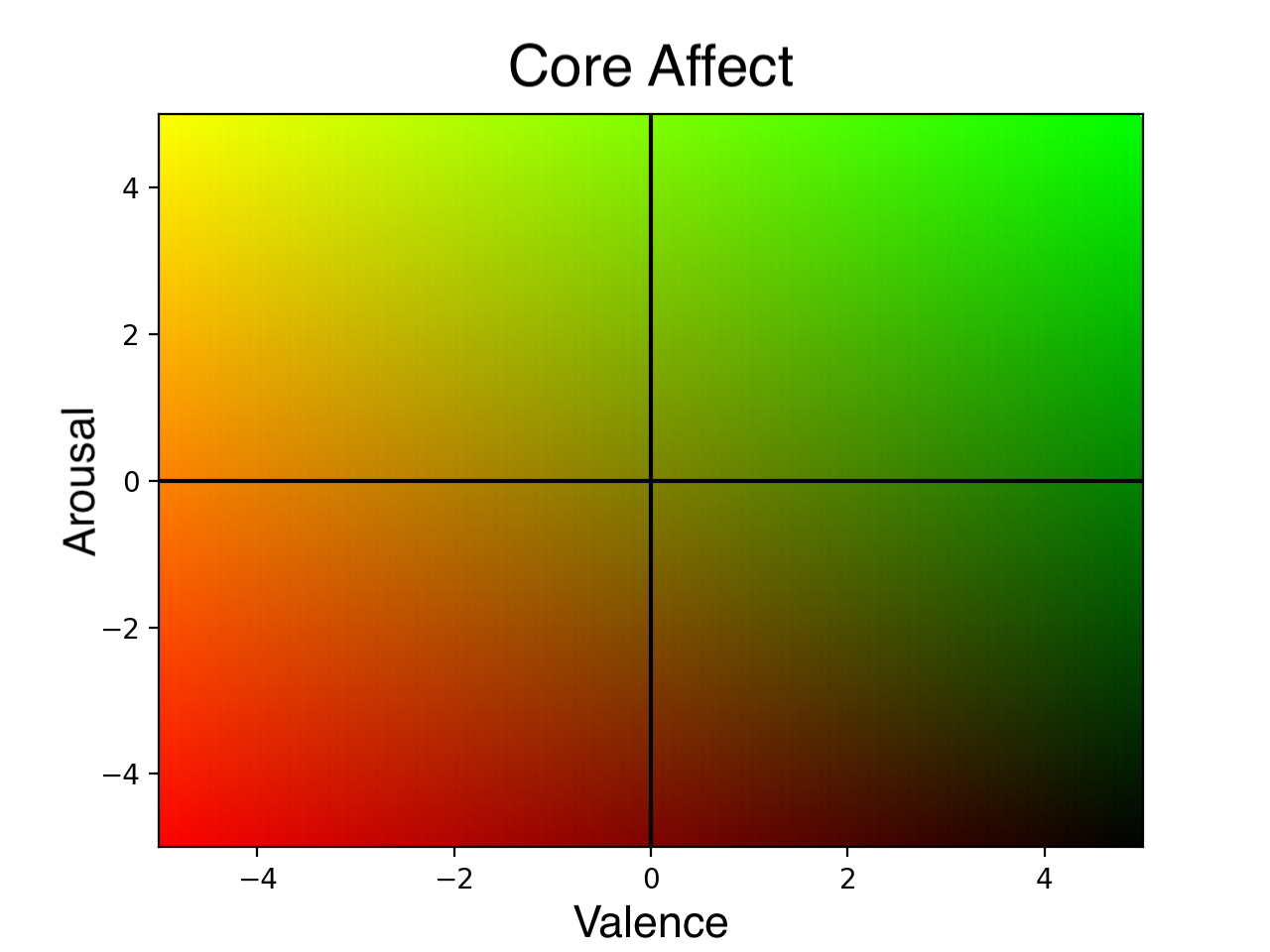}
  \caption{The colour gradation in the two dimensional core affect space, used to represent the artificial affective state of the agent as it traverses the maps. }\label{fig:colourSpace}
\end{figure}

\subsection{Map 1}

\begin{figure*}[!htb]
\includegraphics[width=\linewidth]{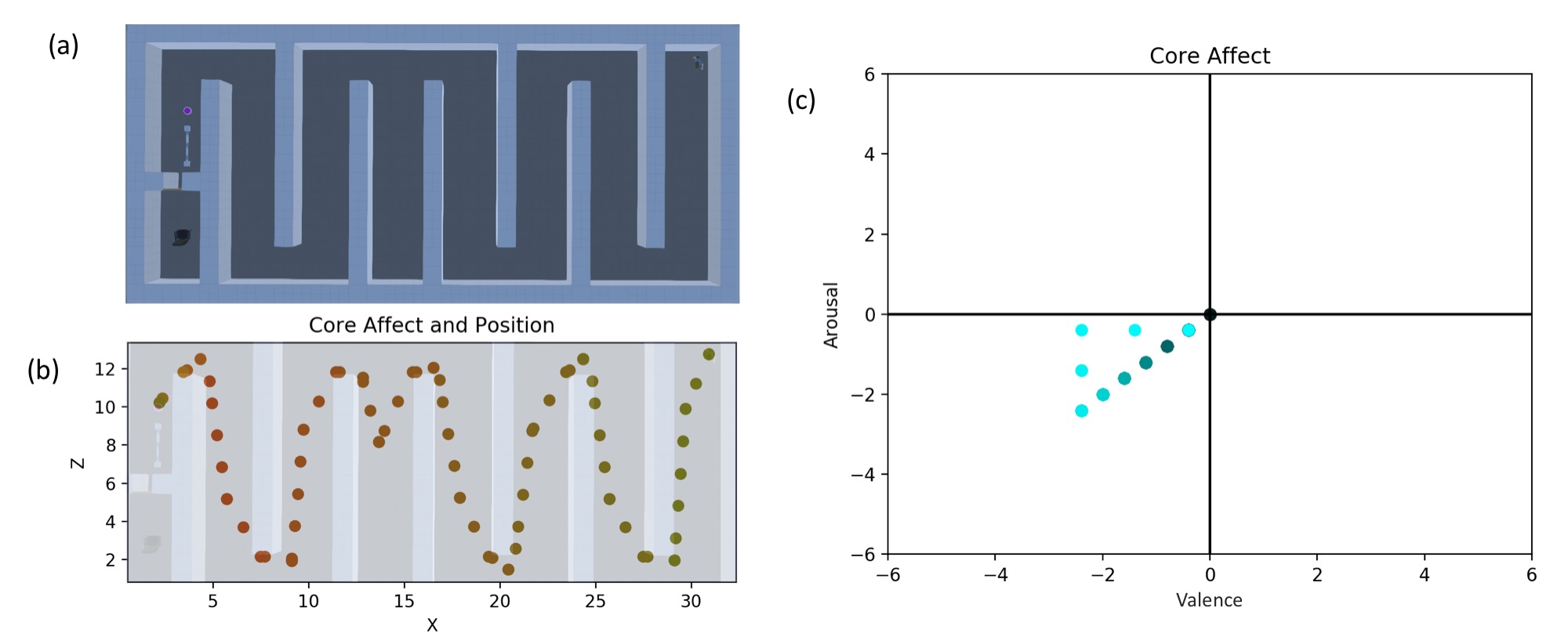}
  \caption{The evolution of the agent's affective state in function of space \textbf{(b)} and time \textbf{(c)} as it traverses Map 1 \textbf{(a)}. The colours in \textbf{(c)} can be mapped to the 2-dimensional Core Affect space using Fig.~\ref{fig:colourSpace}. In \textbf{(c)}, the dots go from black to light blue as time passes. We can see that both affective dimensions turn gradually more negative over time until the agent finally finds the door and opens it, which leads to an increase in both the affective dimensions. The agent's affective state remains in the negative-arousal and negative-valence quadrant throughout the traversal. }\label{fig:Map1}
\end{figure*}

The first Lab Recruits map we will explore is the simplest one (Fig.~\ref{fig:Map1}a). The agent must traverse a maze where it finds no interectables in order to reach a door. When the agent interacts with the button that opens that door, it finds the chair, which is its objective.

The evolution of the agent's affective state in function of time and space can be found on Fig.~\ref{fig:Map1}c and Fig.~\ref{fig:Map1}b, respectively.

The Valence and Arousal dimensions remain negative throughout the agent's traversal of the map. Both Arousal and Valence steadily decrease (Fig.~\ref{fig:Map1}c) until the agent finds the door and the button, where both dimensions finally increase (Fig.~\ref{fig:Map1}b).

\subsection{Map 2}

\begin{figure*}[!htb]
\includegraphics[width=\linewidth]{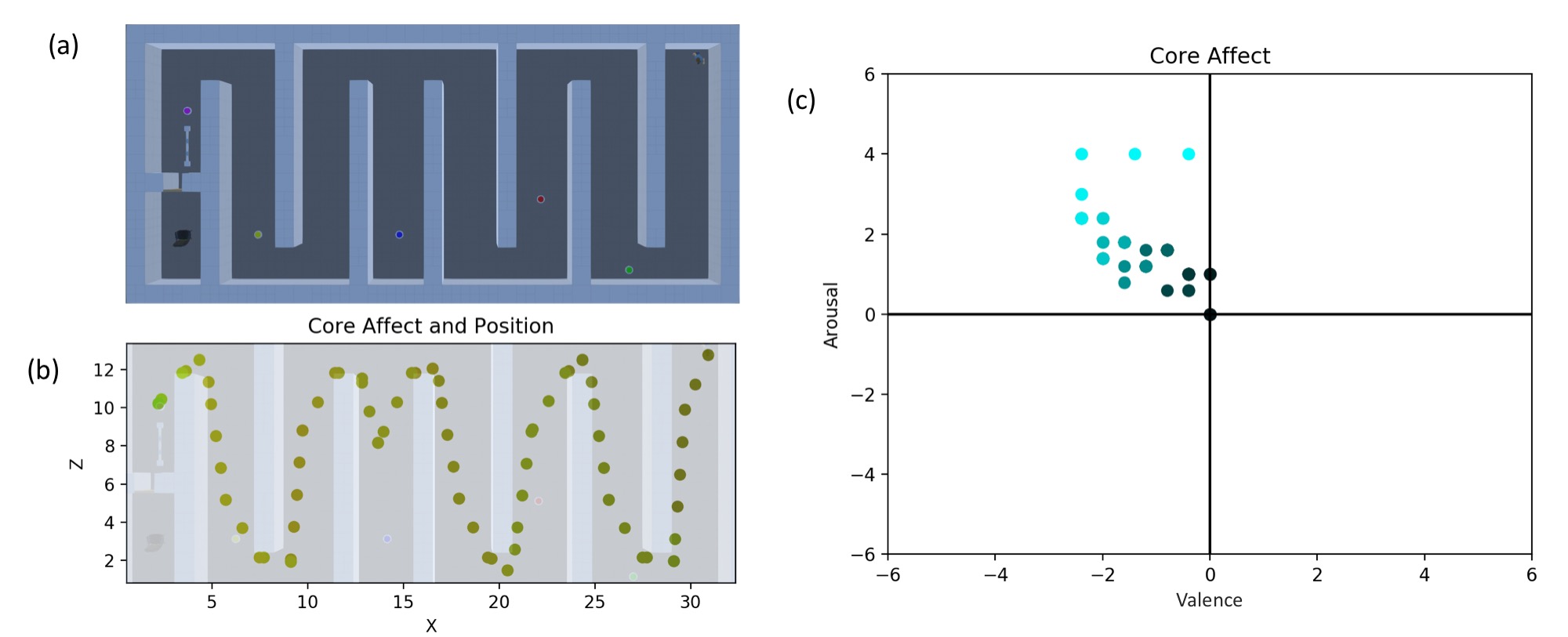}
  \caption{The evolution of the agent's affective state in function of space \textbf{(b)} and time \textbf{(c)} as it traverses Map 2 \textbf{(a)}. The colours in \textbf{(b)} can be mapped to the 2-dimensional Core Affect space using Fig.~\ref{fig:colourSpace}. In \textbf{(c)}, the dots go from black to light blue as time passes. In this map, the arousal dimension increases each time the agent finds a new interactable. The valence dimension, however, steadily decreases until the agent finds the door and opens it, finding the chair. The agent's affective state remains in the positive-arousal and negative-valence quadrant throughout the traversal.}\label{fig:Map2}
\end{figure*}

In the second map, the agent must traverse the same maze as it did in Map 1. However, this time the agent will find interactables on its way to its final objective. These interactables, being buttons not connected to anything, are not relevant to the completion of the agent's objective. 

The evolution of the agent's affective state in function of time and space can be found on Fig.~\ref{fig:Map2}c and Fig.~\ref{fig:Map2}b, respectively.

The Valence dimension of the agent's affective state steadily decreases as it traverses the maze (Fig.~\ref{fig:Map2}c), but this time its Arousal dimension increases each time the agent finds an interactable (Fig.~\ref{fig:Map2}b).

\subsection{Map 3}

\begin{figure*}[!htb]
\includegraphics[width=\linewidth]{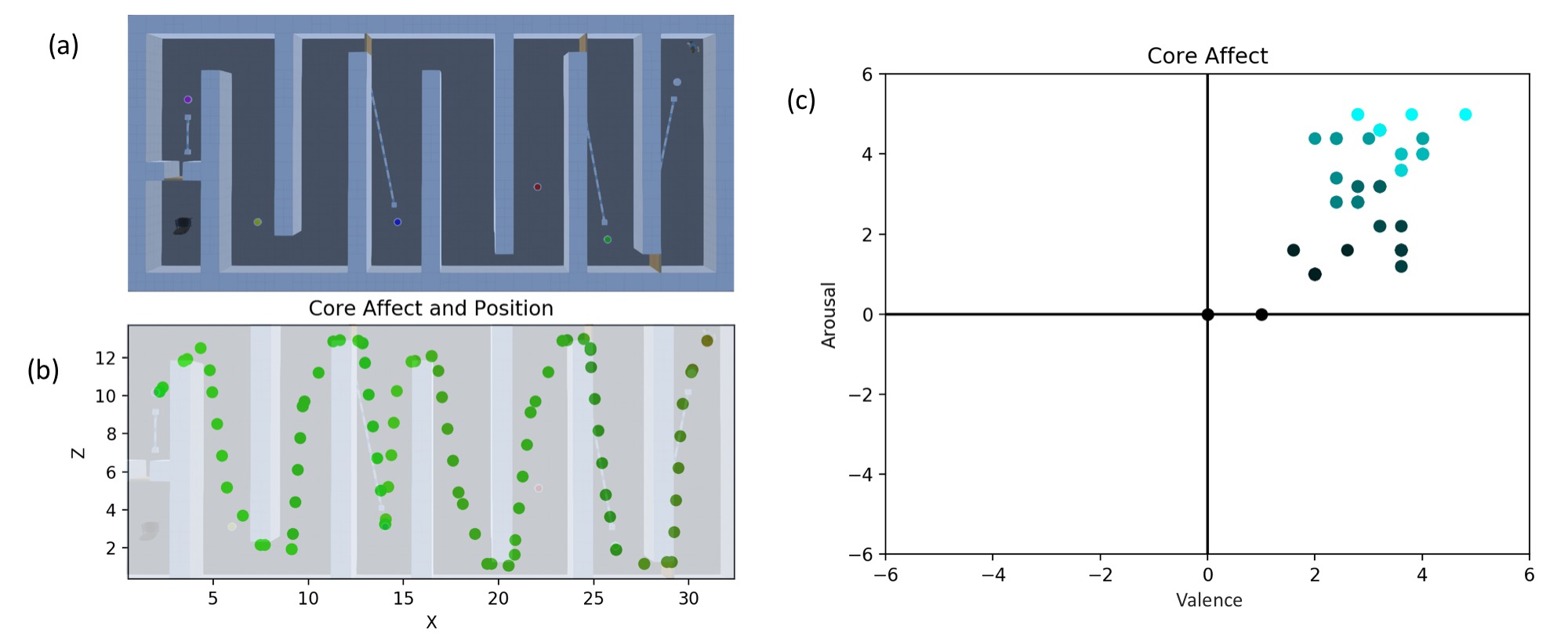}
  \caption{The evolution of the agent's affective state in function of space \textbf{(b)} and time \textbf{(c)} as it traverses Map 3 \textbf{(a)}. The colours in \textbf{(b)} can be mapped to the 2-dimensional Core Affect space using Fig.~\ref{fig:colourSpace}. In \textbf{(c)}, the dots go from black to light blue as time passes. We can see that both affective dimensions increase as the agent traverses the map. Arousal increases whenever the agent finds a new interactable and valence increases whenever the agent is able to accomplish a goal or sub-goal (opening doors). The agent's affective state remains in the positive-arousal and positive-valence quadrant throughout the traversal. }\label{fig:Map3}
\end{figure*}

In Map 3, the agent once again finds itself in a maze. This time, to reach the chair, the agent will not only have to open the final door but 3 other doors that are located throughout the maze. The agent is unable to reach its objective unless the 4 doors are opened. Besides the doors and their corresponding buttons, the agent will find 2 other interactables in the maze. 

The evolution of the agent's affective state in function of time and space can be found on Fig.~\ref{fig:Map3}c and Fig.~\ref{fig:Map3}b, respectively.

Unlike the previous maps, both the Valence and Arousal affective dimensions remain positive throughout the agent's traversal of the map (Fig.~\ref{fig:Map3}c). The agent's Arousal increases whenever it sees an interactable and its Valence increases each time the agent successfully opens a door (Fig.~\ref{fig:Map3}b).

\subsection{Map 4}

\begin{figure*}[!htb]
\includegraphics[width=\linewidth]{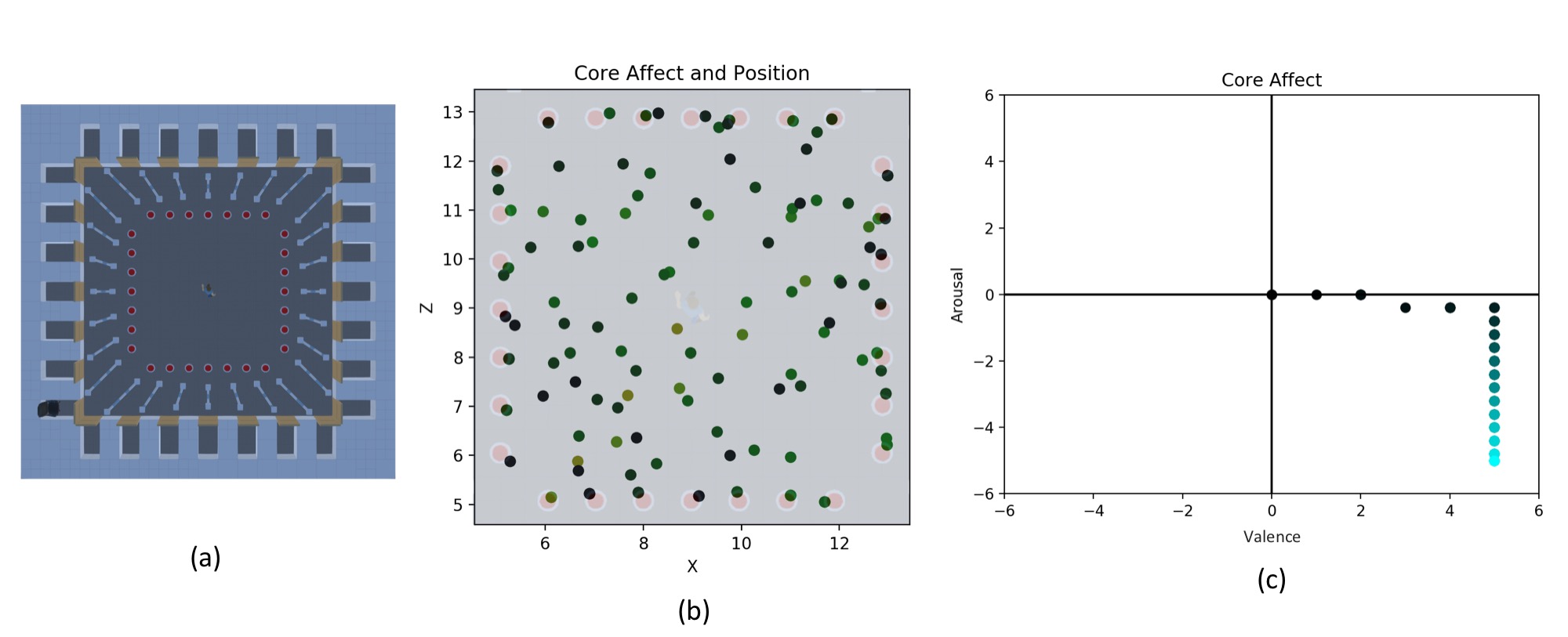}
  \caption{The evolution of the agent's affective state in function of space \textbf{(b)} and time \textbf{(c)} as it traverses Map 4 \textbf{(a)} in the \textbf{worst case scenario}. The colours in \textbf{(b)} can be mapped to the 2-dimensional Core Affect space using Fig.~\ref{fig:colourSpace}. In \textbf{(c)}, the dots go from black to light blue as time passes. This traversal is a chaotic one, with the agent pressing all possible buttons before finally finding the chair. As the agent finds nothing new throughout the traversal, it's arousal dimension steadily decreases, whereas it's valence dimension increases as the agent is able to accomplish sub-objectives (opening doors).}\label{fig:Map4}
\end{figure*}

The fourth map we will analyse is considerably different from the previous 3. This time, the agent is not in a maze but in a room with 28 doors. Each door has a button that opens it and behind only one of the doors, is the chair. The agent can see from the start all the doors and buttons but cannot see the chair before opening the right door.

In this map, the agent's affective experience is very different depending on how ``lucky'' the agent is. In the best case scenario, the agent will open the correct door on the first attempt. On the worst case scenario, the agent will open 27 doors before finally opening the correct one. Because of this, we will here explore both the best and worst case scenarios.

\subsubsection{Best Case Scenario}

In the best case scenario, the first button that the agent activates opens the correct door, leading it to the chair.

In this scenario, the agent's traversal of this map is swift, as the agent must only move from its original position to that of the correct button. As a result, its Arousal dimension doesn't have time to suffer any changes. Its Valence dimension increases as the agent is able to open the door and find the chair.

\subsubsection{Worst Case Scenario}

In the worst case scenario, the agent first opens at random 27 doors before finally opening the correct one, behind which it finds the chair. 

The evolution of the agent's affective state for this scenario in function of time and space can be found on Fig.~\ref{fig:Map4}c and Fig.~\ref{fig:Map4}b, respectively.

Whenever the agent is able to successfully open a door, the agent's Valence dimension increases (Fig.\ref{fig:Map4}b). However, since the agent could see all the buttons and doors since the beginning, there is nothing to increase the agent's Arousal, which steadily decreases with time (Fig~\ref{fig:Map4}c).

\section{Conclusion}
\label{sec:conclusions}

Our agent had considerably different affective responses when traversing each of the 4 different maps that we have tested.

On Map 1 (Fig.~\ref{fig:Map1}), the absence of any interactables throughout the map traversal led the agent's affective state to be on the negative-arousal and negative-valence quadrant during the entirety of the simulation. On Map 2 (Fig.~\ref{fig:Map2}), interactables were scattered throughout the maze, being, however, not relevant to the completion of the agent's objective. This led the agent's affective state to remain on the positive-arousal and negative-valence quadrant. On Map 3 (Fig.~\ref{fig:Map3}), not only does the agent find interactables but also doors that it needs to open in order to accomplish its objective. Both these things make the agent's affective state remain on the positive-arousal and positive-valence quadrant. Finally, on Map 4 (Fig.~\ref{fig:Map4}), the nature of the scenario made it so that the agent would have very different traversals depending on which doors it chose to open. Because of this, we decided to study both the best and the worst case scenario. On the best case scenario, the agent quickly finds the chair, having an increase on the valence dimension of its affective state as a result. On the worst case scenario, the agent continuously opens doors that it had already found and that do not lead to its objective, increasing its valence dimension but making the agent's arousal steadily decrease. In this case, the agent's affective state remains in the negative-arousal and positive-valence quadrant.

We can thus see how different maps arouse different affective states on the test agent. Not only that, but the conclusions we can gather from the evolution of affective states are in accordance with what would be expected. It is not surprising that a map that consists of a monotonous maze will give a player neither arousal nor much valence whereas a maze with challenges that the player can solve and a number of objects to interact with will do the opposite. In maps of greater magnitude, which might have several sections with different densities of challenges and objects, it is not as trivial to understand how the affective state of a player will evolve. The information this approach provides can help developers to pinpoint exactly where changes need to be made in order to arouse in the players the desired affective states.

The very different results found in Map 4 (Fig~\ref{fig:Map4}a) for different traversals of the agent also shows that UX is not only scenario dependent. There will be variability of UX depending on the user's traversal of the system. Running the same scenario several times using stochastic agents or agents with different personas can help designers achieve UX robustness. A design goal could then be a specification of the percentage of agents that would have an UX that fit inside certain parameters.

This paper aims to be a proof of concept, showing that UX testing agents are able to provide information that can be highly useful for both testing and designing. The fact such agents can show UX related information at precise moments and locations also allows them to relay information that would be very difficult to obtain with real users unless physiological methods were employed. Enquiring a user about perceived experience while playing a game will inherently alter the user experience, but that is not the case with UX testing agents.

\section{Discussion}
\label{sec:discussion}

After seeing the examples here presented, the reader might have the following questions: ``But how accurate is this information? How can I be sure the information the agents conveys does indeed reflect how a real user would feel and experience the system?''. Those are very good questions, which we are currently working on to be able to answer. With this paper, our goal was to show how one could employ agents in order to test UX and explore the type of information such agents could provide. We believe our results show that agents could help automate UX testing and make it an integral part of the development process without having bottleneck concerns. In the following paragraphs, we will discuss ways in which this UX testing agents could be improved to be made more accurate and cover different kind of UX tests. 

To ensure the results do indeed correlate with how users experience the system, user testing could be employed in order to fine tune the model to faithfully represent the affective changes users experience. This testing with real users could be done only once in order to attune the UX testing agents to the system under test, allowing the developed agents to be used automatically and without requiring further user testing for the remainder of the system's life-cycle.

This ``tuning'' process could be done, for example, using machine learning and user testing based on physiological methods \cite{callejas2017emotion,moon2016implicit}. Users could be asked to interact with the system under test as both physiological measurements and game events are recorded. The game events, like player location and object interaction, could then be used as input for the machine learning model and an interpretation of the physiological measurements under the emotional model chosen could be the desired output. We could then fine tune the model parameters and the model itself to ensure they accurately represent how a real user experiences the system.

In order to have access to full emotional states instead of affective states, the model could be expanded to have an extra dimension, Dominance, and modified in order to be in accordance with the PAD model of emotions \cite{russell1977evidence}. This model claims emotions can be characterised in a 3-dimensional state, the dimensions being Pleasure, Arousal and Dominance. The pleasure dimension of the PAD model corresponds to the valence dimension used for the core affect model.


The fact that different traversals of the same map lead to different affective results exposes the importance of developing agents display interaction traces that resemble human behaviour. 
If the agents make choices that a user would never do, then the UX information the agents convey might prove to be of no importance. It might also be of no interest to try and program agents to exploit systems in order to find ways of interacting which lead to a negative UX, as the agent might, in the example of a game, decide to continuously walk against a wall and proceed to claim the game can be very uneventful. We thus defend that the development of UX testing agents will not only require well tuned models to assess UX, but also agents that behave similarly to human users.

Finally, we here present a preliminary approach to test a very simple game, but this approach could be modified to test a number of other systems. The objects used in the 4 scenarios here presented could be abstracted to represent other types of objects that are regularly found in games. The door with button can be abstracted as a NPC or interactable which, when interacted with, allows the player to progress towards its objective. The simple button can be abstracted as a NPC or other interactables that are present in games but which are not directly relevant to the completion of the player's objective. The chair can be abstracted as any item that a player must find or a specific location/state that must be reached.

Furthermore, this approach should not be interpreted as game-specific. Agents with a similar architecture could be used, for example, to test the interface of a store, with the valence and arousal dimensions being modelled based on the user finding an item of interest or being able to accomplish a successful purchase or using a coupon. We believe this approach could be used with any system where events that affect the core affect state of an user can be, to a certain degree, identified. This event identification could even be done automatically, using machine learning models and physiological measurements.

\bibliographystyle{IEEEtran}
\bibliography{conference_101719}

\end{document}